\def\year{2021}\relax
\documentclass[letterpaper]{article} 
\usepackage{aaai21}  
\usepackage{times}  
\usepackage{helvet} 
\usepackage{courier}  
\usepackage[hyphens]{url}  
\usepackage{graphicx} 
\usepackage{natbib}  
\usepackage{caption} 
\usepackage[euler]{textgreek}
\usepackage[disable]{todonotes}
\title{Retraining DistilBERT for a Voice Shopping Assistant by Using Universal Dependencies}
\author{
    Pratik Jayarao\textsuperscript{\rm 1},
    Arpit Sharma\textsuperscript{\rm 1}  \\
 
}
\affiliations{
    \textsuperscript{\rm 1}Walmart Labs\\
    Sunnyvale, California

    pratik.jayarao@walmartlabs.com, arpit.sharma@walmartlabs.com

}

\begin{document}

\maketitle

\begin{abstract}
In this work, we retrained the distilled BERT language model for Walmart's voice shopping assistant on retail domain-specific data. We also injected universal syntactic dependencies to improve the performance of the model further. The Natural Language Understanding (NLU) components of the voice assistants available today are heavily dependent on language models for various tasks. The generic language models such as BERT and RoBERTa are useful for domain-independent assistants but have limitations when they cater to a specific domain. For example, in the shopping domain, the token ‘horizon’ means a brand instead of its literal meaning. Generic models are not able to capture such subtleties. So, in this work, we retrained a distilled version of the BERT language model on retail domain-specific data for Walmart's voice shopping assistant. We also included universal dependency-based features in the retraining process further to improve the performance of the model on downstream tasks. We evaluated the performance of the retrained language model on four downstream tasks, including intent-entity detection, sentiment analysis, voice title shortening and proactive intent suggestion. We observed an increase in the performance of all the downstream tasks of up to 1.31\% on average.
\end{abstract}
\section{Introduction}
\todo{----1. advancements in voice assistants are allowing them to be used in domains such as retail especially in these difficult times of COVID when different modes of online shopping are a lot safer than in-store 2. voice assistants have NLU component which currently heavily relies on language models (LMs in intent and entity models) 3. using general domain language models in retail domain has shortcomings such as the embedding of words which mean different in a specific domain than general (e.g., 'great' and 'value' in 'great value') 4. So, in this work we retrained BERT on retail specific data so that it can be used in the NLU component of Walmart's voice shopping assistant}

In recent years, there have been various technological advancements in the field of NLP systems such as voice assistants, allowing them to be used in domains such as retail. Such assistants have become even more important due to COVID-19, helping users adopt different modes of shopping which are safer as compared to in-store shopping. These voice assistants, along with shopping, help users navigate through various tasks such as customer support, search and product tracking. A voice assistant is driven by an underlying NLU engine consisting of various Natural Language Processing models such as Intent Detection, Named Entity Recognition, Sentiment Analysis, Title Compression and Contextual Intent Detection \cite{zhang2020recent, ojamaa2015sentiment}. These models heavily rely on Language Models such as ELMO \cite{peters-etal-2018-deep}, BERT \cite{DBLP:journals/corr/abs-1810-04805}, RoBERTa \cite{DBLP:journals/corr/abs-1907-11692}, and Albert \cite{lan2019albert}. However, these models, trained solely on generic English data, have their shortcomings. Many domain-specific tokens diverge from the meaning associated with them in the general English domain. For example, the token \textit{horizon} widely differs from its usage in the public domain to more of a brand name in the retail space. \todo{Not sure of this example, check with Arpit Sir} Additionally, the queries received by voice assistants are comparatively shorter, have grammatical inconsistencies and other idiosyncrasies. Thus in this work, we retrain the DistilBERT model on a mixture of Retail data and Chat logs, allowing it to be integrated into the NLU engine of Walmart's voice assistant.
\section{Background}
\todo{----------a) What is BERT and How is BERT trained (with loss equation) b) What is DistilBERT and How is DistilBERT trained (using knowledge distillation) and loss equation c) refer to architecture diagrams also here}.

\subsection{BERT} 
BERT (Bidirectional Encoder Representation from Transformers) is used to generate deep contextual embeddings by pretraining it on large-sized corpora. The architecture primarily comprises of multiple blocks of Transformer Encoders \cite{DBLP:journals/corr/VaswaniSPUJGKP17} stacked upon each other. BERT overcomes shortcomings of a solely unidirectional model such as GPT \cite{radford2018improving} by leveraging both, the right and the left context in all layers to generate these representations. They achieve this by introducing a new joint pretraining objective comprising of Masked Language Modeling and Next Sentence Prediction.

These two objectives help BERT to conduct the pretrainng process on extremely large sized corpora and generate rich contextual embeddings. Once the pretraining is completed, the model is further fine-tuned on various downstream tasks.

\subsection{DistilBERT}
Since the introduction of GPT and BERT there has been a slew of highly parameterized transformer-based language models such as RoBERTa \cite{DBLP:journals/corr/abs-1907-11692}, Albert \cite{lan2019albert} and XLNET \cite{yang2019xlnet}. However, there have been studies such as the DistilBERT \cite{sanh2019distilbert}, TinyBERT \cite{jiao2019tinybert, turc2019well} and \citet{liu2020fastbert}  which have reduced parameter size, lowering training and inference time. DistilBERT is trained by utilizing the process of knowledge distillation in which a relatively condensed model (student) is trained to match the functioning of a larger model (teacher). The student, instead of using the cross-entropy loss with the gold standard labels, is trained with a distillation loss over the soft target distribution of the teacher. Equation~\ref{eq:1} represents the loss function for DistilBERT.
\begin{equation} \label{eq:1}
L_{ce} = -\sum t * \log(s)
\end{equation}
Where {\textit{t}} and {\textit{s}}  are teacher and student probability estimations respectively.
These distilled versions of models have been able to achieve relatively good results on multiple downstream tasks such as SQUAD 2.0 \cite{rajpurkar2018know}, CoLA \cite{warstadt2019neural}, STS \cite{cer-etal-2017-semeval}, Quora Question Pairs \cite{DBLP:journals/corr/abs-1907-01041} and MNLI \cite{williams2017broad}.

\subsection{Dependency-Based Word Embeddings}
\citet{levy2014dependency} proposed a generalized SKIP-GRAM model. Instead of using the standard context, they implemented syntactic contexts derived from dependency parse trees. The authors parsed sentences and generated word contexts for a target word \textit{w} with modifiers m\textsubscript{1},.,.,..m\textsubscript{k} and head \textit{h}. Furthermore, they considered contexts \textit{(m\textsubscript{1}, lbl\textsubscript{1}).....(m\textsubscript{1}, lbl\textsubscript{1}), (h,lbl\textsubscript{h}\textsuperscript{1})} where \textit{lbl} is the form of the dependency between the head and the modifier, whilst \textit{lbl\textsuperscript{1}} is used for marking the inverse of this relationship. The authors trained the model on English Wikipedia, with dependencies labelled using the implementation of the parser described in \cite{goldberg-nivre-2012-dynamic}.  This lead to a vocabulary of about 175,000 words, with approximately 900,000 different contexts. In their experiments, they showcased that the earlier SKIP-GRAM models generated more topic-oriented embeddings, whilst incorporating dependency-based contexts encoded more functional characteristics. We use the publicly available dependency-based embeddings released by the authors \footnote{\url{https://levyomer.wordpress.com/2014/04/25/dependency-based-word-embeddings/}} for our experiments. 

\section{Our Approach}
\subsection{Retrained DistilBERT (ReDBERT)}
In this section we will provide a detailed overview of our approach to retrain DistilBERT on retail data. 
\todo{-------Reiterate the goal -\textgreater Building a BERT like language model for retail domain such that it can help in the NLU tasks of a shopping assistant. Since it is to be used in a real world application where inference time plays an important role, we retrained the knowledge distilled version of BERT in this work. Expand on this..... 
}
\subsubsection{Contextual Representations}
 We take an input sentence and tokenize it using a word piece tokenizer. Let X \straightepsilon (x\textsubscript{CLS},x\textsubscript{1}, x\textsubscript{2}..x\textsubscript{T}) be the tokenized input to the model. We then pass these tokens to \textit{DistilBert(.)} in Equation~\ref{distil} which generates corresponding contextual based hidden representations H \straightepsilon  (h\textsubscript{CLS},h\textsubscript{1}, h\textsubscript{2}..h\textsubscript{T}). Where h\textsubscript{CLS} gives a representation for the entire input sequence and h\textsubscript{1} through to h\textsubscript{T} are hidden representations for individual tokens. 
\begin{equation}
\label{distil}
H = DistilBert(X)    
\end{equation}

\subsubsection{Retraining}
One of the primary objectives of the study is to build a BERT like language model for the retail domain to assist the NLU tasks carried out by the shopping assistant. Since the model has to be used in a real-world application, where, inference time plays an important role, we retrain the distilled version of BERT instead of its larger counterparts. We utilize the standard DistilBERT architecture described in \citet{turc2019well} and train a two-layer and a four-layer model with 768 as the hidden representation size.
We initialize the DistilBERT model with pretrained weights \footnote{Pretrained Weights BERT - https://github.com/google-research/bert}. This helps us transfer fundamental English knowledge to our model from the pretained DistilBERT model. However, we do not employ the teacher-student paradigm with knowledge distillation loss in Equation~\ref{eq:1} for the further retraining process on retail data.  Utilizing knowledge distillation would require us to go through the process of retraining the larger versions of BERT (teacher) on the retail corpus first. However, the computationally expensive nature of training multi-layered transformer-based models like large BERT requires clusters of GPUs / TPUs. Since we cater our study towards researchers with low resources, we decided to carry out the retraining process utilizing the original BERT training objectives Next Sentence Prediction and Masked Language Prediction.

\textbf{Next Sentence Prediction (NSP)} is the task of predicting if two segments presented to the model follow each other. We pass the hidden representation obtained in Equation~\ref{distil} through Equation ~\ref{eq:pnsp} to obtain the Probability P\textsubscript{NSP} of the next segment following the earlier one. We train the model by optimizing the standard cross-entropy loss presented in Equation ~\ref{eq:lnsp}.

\begin{equation}
\label{eq:pnsp}
P_{NSP} = softmax( h_{CLS} * W_{NSP} + B_{NSP}))    
\end{equation}

\begin{equation} 
\label{eq:lnsp}
L_{NSP} = -\sum y_{NSP} * \log(P_{NSP})
\end{equation}

\textbf{Masked Language Modeling (MLM)} is the task of predicting randomly masked tokens in a sentence. We utilize the token representations obtained in Equation~\ref{distil} and generate probability of masked tokens P\textsubscript{MLM} through Equation~\ref{eq:pmlm}. We furthermore train our model using the standard cross-entropy as mentioned in Equation ~\ref{eq:lmlm}.

\begin{equation}
\label{eq:pmlm}
P_{MLM} = softmax(h * W_{MLM} + B_{MLM})
\end{equation}

\begin{equation}
\label{eq:lmlm}
L_{MLM} = -\sum\sum y_{MLM} * \log(P_{MLM})
\end{equation}

The formulation mentioned above helps us circumvent training the large BERT model and allows us to conduct our study with limited infrastructure. Additionally, the two-layer and the four-layer DistilBERT models comprise of 39.2, and 53.4 million parameters whilst the original BERT large model contains  340 million parameters. This reduction in parameters helps in decreasing the training time significantly and makes it feasible to complete the experiments with low resources. Lastly, we abstain from making any alterations to the vocabulary and employ the \textit{BERT base vocab} for training our model.


\subsection{Dependency Embeddings Injected DistilBERT (DeDBERT)}
\todo{------Here mention about the need/intuition of dependency injection. Also refer to the architecture diagram as well as add the loss equation here}

The recently released large transformer-based encoders which achieve state of the art performance on various tasks comprise of a large number of parameters. Since we employ a smaller model, it leads to a loss in the abilities of the model \cite{turc2019well}. To correct for this, we search for various techniques to aid the model's performance without exponentially increasing the number of parameters.

\begin{figure*}[t!]
  \centering
    \includegraphics[width=14cm, height=4.5cm]{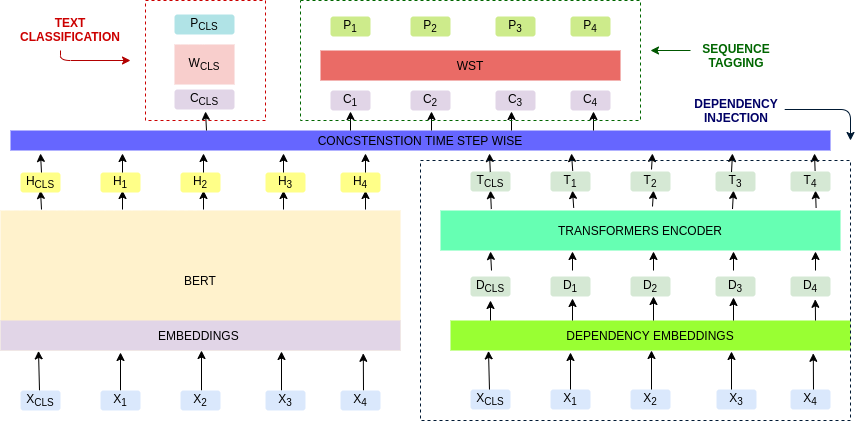}
  \caption{DistilBERT with Dependency Injection}
  \label{fig:dependency}
  
\end{figure*}

\citet{komninos-manandhar-2016-dependency} used dependency-based word embeddings and observed an increase in the performance of tasks which were driven by the syntax of the input text. These results achieved by the authors motivated us to experiment with dependency-based word embeddings for our syntax dependent downstream tasks. We hypothesize that this form of explicit introduction of information to the model would help the model perform better. Additionally, since the dependency embeddings do not require any form of hand labelled data, it naturally lends itself to the retraining (which comprises of large unlabelled corpora). Moreover, no human expertise is required to incorporate the same in the downstream tasks. We thus inject dependency-based word embeddings to provide external syntactic context to the model. We propose a novel technique for injecting dependency-based embeddings into the two-layered DistilBERT model in our experiments using a single-layered transformer encoder. The transformer encoder follows the standard architecture described in \citet{DBLP:journals/corr/VaswaniSPUJGKP17}. An example of the injection process, along with the extended Text classification and Sequence tagging task is showcased in Figure \ref{fig:dependency}.

We obtain a pretrained Dependency Word Embeddings matrix DW \straightepsilon R\textsubscript{Vx300} where V is the vocabulary size. We look up the corresponding embeddings for each token present in X and then pass it through a \textit{Transformer(.)} to obtain the transformer representations T \straightepsilon  (t\textsubscript{CLS}, t\textsubscript{1}, t\textsubscript{2} .. t\textsubscript{T}). These representations are then concatenated at every time step with X to generate C \straightepsilon  (c\textsubscript{CLS}, c\textsubscript{1}, c\textsubscript{2} .. c\textsubscript{T}). This process is summarized in Equations~\ref{eq:t1}, ~\ref{eq:t2} and ~\ref{eq:t3}.

\begin{equation}
D = lookup(X, DW)
\label{eq:t1}
\end{equation}
\begin{equation}
T = Transformer(D)
\label{eq:t2}
\end{equation}
\begin{equation}
C = Concatenation(T, H)
\label{eq:t3}
\end{equation}

We use C instead of H to further to carry the retraining process as described in the previous section. 
This architecture helps us keep the innate structure of the DistilBERT model and also takes advantage of the parallelism on GPU lent by self-attention architectures. Moreover, it can be directly used for the pretraining/downstream tasks without changing their respective training objectives or data preprocessing steps.

\subsection{Data For Retraining}
\todo{------Mention that this work is done for improving the NLU component of a publicly available voice shopping assistant. The assistant currently supports a subset of all the products sold by the retailer.}
The primary goal of the study is to improve the underlying NLU engine of a publicly available voice shopping assistant. The shopping agent currently supports a subset of all the products sold by the retailer. This leads to the cropping up of these products in the conversations carried out by the agent.

\noindent 
Thus we extract the following two kinds of training data.

\begin{enumerate}
\item \textbf{Product Data: } It consists of the titles and the descriptions of the products that are supported by the Walmart's voice shopping assistant. Examples of the same can be found in Table~\ref{tabe:Data1}.
\item \textbf{Chat Logs from the Voice Shopping Assistant:} The general domain language models are not trained on chat data and our goal is to be able to use the retrained model for the voice assistant. So, we also incorporate the annotated chat data from real users in the retraining process showcased in Table~\ref{tabe:Data2}. \todo{-------Add examples in the same table and link it here}.
\end{enumerate}

\todo{--------Briefly explain what is an instance of data in DistilBERT training. Then - we were able to extract a total of xyz data instances from the above two data sources. Out of those abc were from Product Data and a1b1c1 from Chat Logs. The data was divided into a\% and b\% splits for training and testing respectively.}
\noindent
A single training instance used in the DistilBERT has a fixed number of maximum tokens (Max Sen Length) which is 128. We retrain the model on a small corpus consisting of 750 MB of data, i.e. 25 million instances, with approximately 20\% of data coming from Chat Logs and the rest from Walmart's Product Catalog. We further split the data into a 90-10 training-test split for the retraining process. Despite retraining on such a small corpus we observe a boost in the performance of our downstream tasks.

\begin{table}[htpb]
\centering
\begin{tabular}{ |c||l|  }
 \hline
 \multicolumn{2}{|c|}{Product Data} \\
 \hline
 Product Title& Product Description\\
 \hline
 Apple&   Active Noise Cancellation for \\
 AirPods&  immersive sound. Transparency  \\
 Pro & mode for hearing and connecting\\
 &  with the world.\\
 \hline
\end{tabular}
\caption{Product data examples}
\label{tabe:Data1}
\end{table}
\begin{table}[htpb]
\centering
\begin{tabular}{ |l|  }
 \hline
 \multicolumn{1}{|c|}{User Utterances} \\
 \hline
 Hi, I want to buy Apple air pods \\
 
 I want to buy the Apple Pro version\\
 
 Can you also add a screen guard to cart\\
 
 What is the total?\\
 
 Remove the screen guard from cart\\
 \hline
\end{tabular}
\caption{User utterance examples}
\label{tabe:Data2}

\end{table}

\section{Experiments and Results}

\todo{-------Complying with the objective of improving the NLU components of our voice shopping assistant, we performed an in-vivo evaluation of our approach on various NLU tasks. In the following sections, we first provided the setup used for retraining BERT. Then we provided the brief explanation of each task and the comparison between the vanilla BERT and our retrained BERT (henceforth called ReDBERT)}
Complying to improve the NLU components of Walmart's voice shopping assistant, we perform an in-vivo evaluation of our approach on various NLU tasks. In the following sections, we provide the setup used for retraining DistilBERT.
\subsection{Setup for Retraining}
We initialize DistilBERT with pretrained weights.
We train two versions of DistilBERT. A two-layer and four-layer version with the hidden representation size of 768. Since one of the goals of the study was to perform experiments which are easy to reproduce and achievable with minimal infrastructure, we make use of a single NVIDIA V100 32GB GPU. We deploy early stopping to ensure that the model does not overfit the small corpus. We used Adam optimizer with a constant learning rate of 2e-5 and max length of 128.

\begin{table}[htpb]
\centering
\begin{tabular}{ |c||c|c|c|  }
 \hline
 \multicolumn{4}{|c|}{Retraining Times} \\
 \hline
 Parameter     & 2 Layer & 4 Layer & 2 Layer DeDBERT\\
 \hline
 Batch Size&   600& 300& 256  \\
 Epoch Time& 11& 20& 15 \\
 Total Time& 10 days& 6 Days & 5 days \\
 \hline
\end{tabular}
\caption{Retraining}
\label{hyp:2}

\end{table}

\subsection{Downstream Tasks}
We evaluated the retrained model on four different NLU tasks associated with Walmart's voice shopping assistant. All the downstream tasks we performed our evaluations on fall under the two broad categories. \textbf{1.Text Classification}-(Intent Detection, Contextual Intent Detection, Sentiment Analysis) \textbf{2. Sequence Tagging}-(NER, Title Compression). A brief overview and evaluation of each task is showcased in the sections below.

\subsubsection{Text Classification}
We use the hidden representation for the classification token obtained in Equation~\ref{distil} and generate the probability for the target classes, as mentioned in Equation~\ref{eq:pcls}. We train the model by optimizing the standard cross-entropy loss L\textsubscript{CLS} where y\textsubscript{CLS} is the true label.
\begin{equation}
\label{eq:pcls}
P_{CLS} = softmax( h_{CLS} * W_{CLS} + B_{CLS})
\end{equation}
\begin{equation} \label{eq:1_2}
\label{loss:cls}
L_{CLS} = -\sum y_{CLS} * \log(P_{CLS})
\end{equation}

\subsubsection{Sequence Tagging}
We generate the hidden representations for every input token from Equation~\ref{distil} and use Equation~\ref{eq:pt} to obtain the probability for classes for each token P\textsubscript{t}. We train this model by optimizing the standard cross entropy loss L\textsubscript{ST} where y\textsubscript{ST} is the true label.
\begin{equation}
\label{eq:pt}
P_{ST} = softmax(h * W_{ST} + B_{ST})
\end{equation}
\begin{equation} \label{eq:1_1}
\label{loss:st}
L_{ST} = -\sum\sum y * \log(P_{ST})
\end{equation}

\subsubsection{Joint Intent-Entity Detection}
Intent detection\todo{cite here} is the process of associating a given user utterance to a specific goal. It is a multi-class \textit{Text Classification} task in which the user's utterance is analyzed and assigned a particular intent. For example, let us consider the user utterance \textit{``add milk to my cart''}. The intent of the query could be \textit{`add\_to\_cart'}. Intent recognition plays a crucial role in directing the flow of the conversation towards the goal.

Similar to the Intent detection, Named Entity Recognition (NER)\todo{cite} is a classic NLU task for a shopping assistant. It consists of extracting spans of entities from a given text. It is a standard \textit{Sequence Tagging} task where each token is assigned to a particular label (or entity type). For example, \textit{`Apple'} is a named entity of type \textit{`brand'} in the user utterance \textit{``I want to purchase Apple Airpods Pro''}.
It directly contributes to the success of the dialogue system.

\todo{------Briefly explain how a BERT based joint intent and entity model works}
The retraining process advances the performance of both the above tasks with Intent Detection seeing an increase of 0.622\% and Named Entity Recognition of 0.788\%.
\todo{Think how we can combine all the downstream task architecture diagrams into one}

\subsubsection{Sentiment Analysis}
Sentiment analysis is the task of analyzing if a given text comprises of a positive, negative or a neutral connotation. It is a \textit{Text classification} task, and in real-time, this cue helps the agent understand the underlying tone of the user utterance and respond appropriately. In post conversation analysis, the overall sentiment of a conversation is an essential metric for dialogue systems and various customer satisfaction studies. It also helps in finding flaws at particular points in a failed conversation. For example, we can assess the junctures of a conversation where negative sentiment was present and further investigate the reasons for the same. Our version of the DistilBERT model helped significantly boost the F1 performance of the task by 1.65 \%. 

\subsubsection{Title Compression}
Title compression is the process of obtaining the title of the product from the product description. \todo{-----------Provide details on how and why this is a crucial component of a voice shopping assistant. You can say ``the actual product titles in the catalogue are very long because they consist of size, brand and many other details about a product. For example ..... But providing such long product titles to a user on a voice assistant is not a good idea as it would not be a good user experience. so shortening of titles is needed.''} Products present in Walmart's catalogue, in general, have long titles. They comprise of granular information like the size, brand, tech specifications and other finer details. Providing such verbose titles to the users directly would hurt the user experience, and thus shortening of titles is necessary.

We transform the task from a generative form to an extractive nature by converting it to 1/0 \textit{Sequence Tagging} task. The retrained model improves the performance of the task by 0.296 \%. This model plays an important role in the UX.

\subsubsection{Proactive Intent Detection}
Proactive Intent Detection works on lines similar to that of the Intent detection model and is a \textit{Text Classification} task. However, it takes into account the conversation history along with the latest intent and then proactively predicts the next intent. This predicted intent is then sent across as a suggestion to the user. This task aids a user in reducing the time to complete a given task, along with lowering conversational confusion resulting in more successful conversations. If a user requests to \textit{"remove an item from cart"}. The assistant along with confirmation messages \textit{"Affirmative"} and \textit{"Negative"} will also suggest \textit{"Show my cart"} as an option to the user.\todo{-----Provide an example}. With the retrained DistilBERT model, we see a rise of 0.984 \% in the F1 scores of the model.

Proactive Intent Detection falls under the umbrella of Text Classification. The input X comprises of two parts \textit{X\textsubscript{A} = (x\textsubscript{\textit{}CLS}, x\textsubscript{1}, x\textsubscript{2}, ...x\textsubscript{TA})} and \textit{X\textsubscript{B} = (x\textsubscript{1}, x\textsubscript{2} ...x\textsubscript{TB})} concatenated to form X and are separated by [SEP] token. \textit{X\textsubscript{A}} comprises of the current user utterance and \textit{X\textsubscript{B}} comprises of the previous user utterances. We obtain H from Equation~\ref{distil}. Additionally, we pass the current intent \textit{X\textsubscript{I}} through a neural network \textit{NN(.)} comprising of 2 layers to obtain a representation \textit{I}. We then concatenate \textit{h\textsubscript{CLS}} and \textit{I} to obtain \textit{hi\textsubscript{CLS}}.


\begin{table}[htpb]
\centering
\begin{tabular}{ |c|c|c|  }
 \hline
 \multicolumn{3}{|c|}{Downstream Tasks 2 layer DistilBERT} \\
 \hline
 Metric     & DistilBERT & ReDBERT\\
 \hline
 Intent Detection&   81.091   & \textbf{81.713}\\
 Entity Detection & 91.964 & \textbf{92.752}\\
 Proactive Intent Detection & 78.568 & \textbf{79.552}\\
 Sentiment Analysis & 65.511 & \textbf{67.161}\\
 Title Compression & 88.688 & \textbf{88.984}\\
 \hline
\end{tabular}
    \caption{F1 Scores for Downstream Tasks 2 layer}
\label{table:res1}

\end{table}

\begin{table}[htpb]
\centering
\begin{tabular}{ |c|c|c| }
 \hline
 \multicolumn{3}{|c|}{Downstream Tasks 4 layer DistilBERT} \\
 \hline
 Metric     & DistilBERT & ReDBERT\\
 \hline
 Intent Detection&   77.936   & \textbf{78.289}\\
 Entity Detection & 92.178 & \textbf{92.724}\\
 Contextual Intent Detection & 78.568 & \textbf{79.552}\\
 Sentiment Analysis & 66.006 & \textbf{67.316}\\
 Title Compression & 89.222 & \textbf{89.558}\\
 \hline
\end{tabular}
\caption{F1 Scores for Downstream Tasks 4 layer}
\label{table:res2}

\end{table}

Each downstream task requires a specialized form of intelligence. Whilst some are dependent heavily on semantics, some on syntax and some on both.
We observe an increase in the performance of the model on all the downstream tasks. This holds for both the two and four-layered model. The results have been summarized in Tables~\ref{table:res1} and \ref{table:res2}.

\subsection{Dependency Injection}
\todo{We need more experiments here. We will have to evaluate dependency retraining performance on all the tasks and mention the effect of it in those tasks itself, i.e., in the above sections.}
\subsubsection{Retraining with Dependency Injection}
We retrain a two-layered DistilBERT on retail and chat data along with dependency-based word embeddings. We allow the dependency-based word embeddings used for injection to be fine-tuned along the retraining process. The process for retraining is the same as described in the earlier section.

\subsubsection{Downstream Tasks}
Since dependency-based word embeddings provide syntactic contexts, we evaluate the model on syntactic based sequence tagging tasks such as Named Entity Recognition and Title Compression. The C\textsubscript{CLS} is multiplied with W\textsubscript{CCLS} and with addition of B\textsubscript{CCLS} W\textsubscript{CCLS} \straightepsilon R\textsuperscript{cxn} followed by softmax and argmax to obtain P\textsubscript{CCLS} and similary we obtain P\textsubscript{CST} using W\textsubscript{CST} \straightepsilon R\textsuperscript{cxm} and B\textsubscript{CST} \straightepsilon R\textsuperscript{m}. We use the loss functions showcased in Equations~\ref{loss:cls} and \ref{loss:st} for the text classification and sequence tagging tasks respectively. The results for syntax based downstream tasks have been summarized in Table ~\ref{table:syntax}
\begin{equation}
P_{CCLS} = softmax( c_{CLS} * W_{CCLS} + B_{CCLS})
\end{equation}
\begin{equation}
P_{CST} = softmax(c * W_{CST} + B_{CST})    
\end{equation}

\begin{table*}[htpb]
\centering
\begin{tabular}{ |c|c|c|c|c|  }
 \hline
 \multicolumn{5}{|c|}{Downstream Tasks F1 Scores} \\
 \hline
 Metric     & 2L DistilBERT& 4L DistilBERT & DeDBERT & Retrained DeDBERT\\
 \hline
 Entity Detection & 91.964& 92.178 & 92.238& \textbf{92.369}\\
 Title Compression & 88.688& 89.222 & 90.032&\textbf{90.159}\\
 \hline
 
\end{tabular}
\caption{F1 Scores for syntax based Downstream Tasks}
\label{table:syntax}

\end{table*}

\section{Discussions}
\todo{In this section we can do the following: 1) discuss our findings by using visual aids (graphs and all) 2) present the remaining errors in tasks and their causes and possible solutions}

The retraining process was conducted with only a small-sized corpus. Additionally, we accomplish the retraining process without using large clusters of GPUs and have shorter training duration. Even with the above-stated restrictions, our retrained model outperforms the original DistilBERT model on all downstream tasks. This indicates that the retrained model can capture the distribution of the underlying retail domain data. This showcases that even the usage of a small corpus can help achieve improvement in the domain-specific downstream tasks. We also used the DeDBERT architecture for the retraining process and observe a similar increase in the performance on the syntax tasks. 

\begin{figure}

  \centering
    \includegraphics[width=5cm, height=5cm]{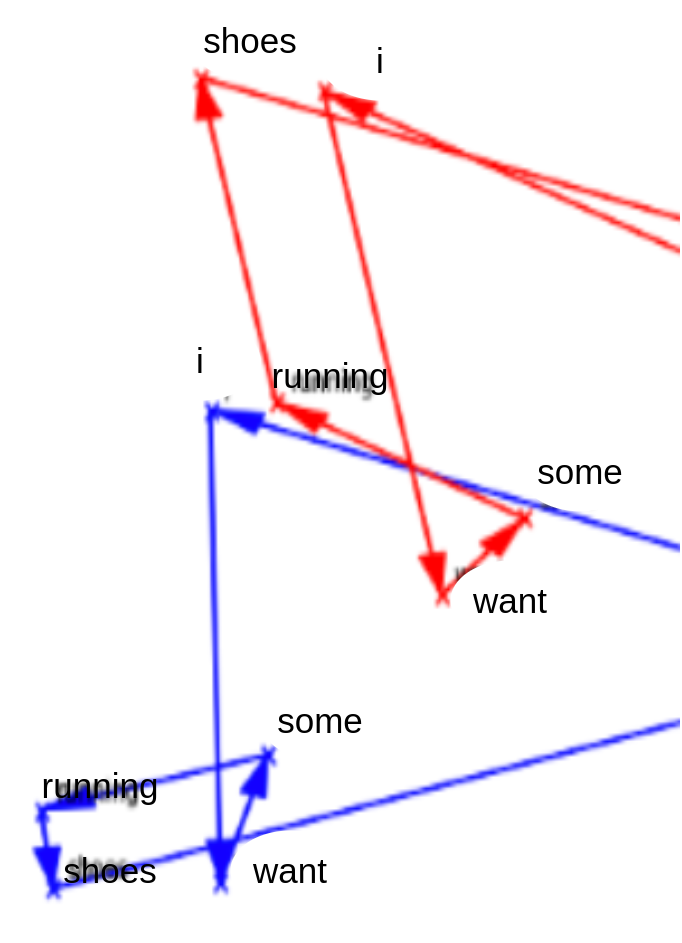}
  \caption{Visualization for \textbf{\textit{i want to buy some running shoes}}. Red - Original DistilBERT. Blue - ReDBERT}
  \label{fig:viz1}
  
\end{figure}

\begin{figure}

  \centering
    \includegraphics[width=5cm, height=5cm]{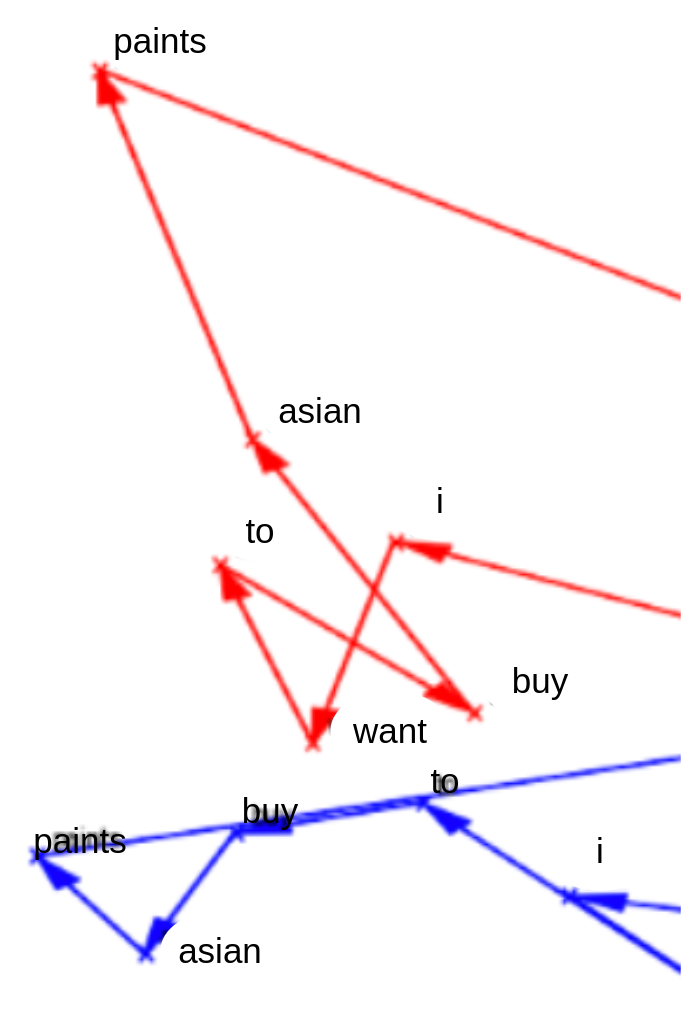}
  \caption{Visualization for \textbf{\textit{i want to buy asian paints}}. Red - Original DistilBERT. Blue - ReDBERT}
  \label{fig:viz2}
  
\end{figure}

In order to analyze the effect of the retraining process, we compare the contextual embeddings generated by the two-layered DistilBERT vs the two-layered retrained DistilBERT. We generated the embeddings by both the above-stated models and projected them into two-dimensional space using Principal Component Analysis (PCA). An illustration of the same can be seen in Figures~\ref{fig:viz1} and ~\ref{fig:viz2}. We measure the distance between tokens using Euclidean Distance.  The diagrams clearly showcase that there have been notable changes in the representations generated by the retrained model.  The above stated changes in the embeddings occur for multiple reasons. Tokens which in the general domain stand for different concepts are realigned according to the domain specific corpus (especially for brand names). For example tokens \textit{asian} and \textit{paints} in Figure~\ref{fig:viz2} are placed close to each other (1.769 units) by the retrained DistilBERT model compared to the original DistilBERT model (4.588 units). This realignment results from the high frequency of the occurrences of the above stated tokens in their domain specific contexts. The model goes through the corpus adjusting to the new usages of the tokens in their peculiar contexts and is able to achieve the above.  
Secondly, the model also places highly queried product n-grams such as \textit{running} \textit{shoes} quite close to each other in Figure~\ref{fig:viz1} (0.873 units) whilst the original model places the tokens at a distance of (3.317 units). The model observes the tokens in the above n-grams placed in close proximity to each other in the training corpus. Many such other subtle changes caused due to the retraining process but difficult to infer visually also lead to the generation of fine grained contextual embeddings catering to retail.

\subsection{Dependency Injection}
We introduced a novel injection technique to incorporate dependency-based word embeddings into our model allows the model to obtain more syntactic information for the input tokens. The two, four layered DistilBERT consists of 39.2 and 53.4 million parameters whilst the DeDBERT comprises of 51.2 million parameters. However, the DeDBERT model outperforms the two-layered DistilBERT and in particular the four-layered DistilBERT model. Despite, lower number of parameters compared to the four-layered model the DeDBERT model performs better showcasing that the model does not just gain performance due to the increase in the parameters from the transformer but can benefit from the explicit knowledge provided by the dependency-based embeddings. Moreover, the DeDBERT model converges faster than its counterparts with lower inference time compared to four-layered model.

\section{Related Work}
In recent years there has been a wave of transformer \cite{DBLP:journals/corr/VaswaniSPUJGKP17} based language models such as GPT \cite{brown2020language}, BERT \cite{DBLP:journals/corr/abs-1810-04805}, RoBERTa \cite{DBLP:journals/corr/abs-1907-11692} and \citet{lan2019albert}. They have achieved state-of-the-art results on a wide variety of downstream tasks such as SQUAD 2.0 \cite{rajpurkar2018know}, CoLA \cite{warstadt2019neural}, STS \cite{cer-etal-2017-semeval}, Quora Question Pairs \cite{DBLP:journals/corr/abs-1907-01041}, MNLI \cite{williams2017broad} \label{Academic:1}. However, these models are trained on generic English corpus and do not generalize well on domain specific data. Consequently, there have been attempts to retrain transformer based models on domain-specific corpora, showcasing in an improvement in the performance on these downstream tasks.
 
 \citet{DBLP:journals/corr/abs-1901-08746} retrained BERT on a large bio-medical corpus. They initialized BERT with pre-trained weights and trained it till convergence. They did not alter the original BERT vocabulary for the domain-specific training process. The retrained model is called BIO-BERT. \citet{DBLP:journals/corr/abs-1903-10676} retrained BERT on the scientific corpus. The retrained model is called SCI-BERT. They primarily trained two versions of BERT. BERT initialized with pre-trained weights and BERT's original vocabulary. BERT trained from scratch with new domain-specific vocabulary (SCI-VOCAB).
 Both BIO-BERT and SCI-BERT showcased an increase in the accuracy of the downstream, domain-specific tasks. The majority of improvement in the performance was noted due to the retraining process rather than the incorporation of new vocabulary. Similar attempts have also been made by \citet{DBLP:journals/corr/abs-1904-03323} on clinical data.
\section{Conclusion}
We retrained two and four-layered versions of DistilBERT on retail and chat data. We utilized a small-sized corpus and limited infrastructure to generate these retail-specific contextual embeddings. We evaluated the retrained model on multiple downstream tasks deployed in goal-oriented dialogue systems and observed an increase in performance. Additionally, we also conducted a visual evaluation of the embeddings generated by the retrained DistilBERT model. 
Moreover, we proposed a novel architecture incorporating dependency-based word embeddings into the DistilBERT model and saw an improvement in the performance on sequence tagging downstream tasks.
\section{ Acknowledgments}
We would like to acknowledge the support of our colleagues in the Emerging Technology team at WalmartLabs, Sunnyvale, CA office.

\bibliography{aaai2021}
\bibstyle{aaai21}

\end{document}